\documentclass[letterpaper, 10 pt, journal, twoside]{IEEEtran}  

\IEEEoverridecommandlockouts                              

\usepackage{graphicx} 
\usepackage{amsmath} 
\usepackage{amssymb}  
\usepackage{subfigure}
\usepackage{hyperref}
\usepackage[noadjust]{cite}

\newcommand\Bstrut{\rule[-0.9ex]{0pt}{0pt}}

\title{
GRASPA 1.0: GRASPA is a Robot Arm gra{S}ping Performance benchm{A}rk
}

\author{Fabrizio Bottarel$^{1,2,*}$, Giulia Vezzani$^{1,*}$, Ugo Pattacini$^1$, and Lorenzo Natale$^1$%
	\thanks{To cite this work, please refer to F. Bottarel, G. Vezzani, U. Pattacini and L. Natale, "GRASPA 1.0: GRASPA is a Robot Arm graSping Performance BenchmArk," in IEEE Robotics and Automation Letters, vol. 5, no. 2, pp. 836-843, April 2020.}
	\thanks{Digital Object Identifier (DOI): 10.1109/LRA.2020.2965865}
	\thanks{Manuscript received: August 13, 2019; Revised: November 17, 2019;
		Accepted: December 12, 2019.}
	\thanks{This paper was recommended for publication by Editor Han Ding upon
		evaluation of the Associate Editor and Reviewers' comments.}
	\thanks{This work was supported by the European H2020 project No. 730994 (TERRINet) and ERA-NET CHIST-ERA call 2017 project HEAP.}
	\thanks{*Equal contribution.}%
	\thanks{$^{1}$Fabrizio Bottarel, Giulia Vezzani, Ugo Pattacini and Lorenzo Natale are with
	Istituto Italiano di Tecnologia, via San Quirico 19D, Genova, Italy
	        {\tt\small name.surname@iit.it}.}%
	\thanks{$^{2}$ Fabrizio Bottarel is also with Department of Informatics, Bioengineering, Robotics and Systems Engineering, Universit\`a di Genova, Genova, Italy.}%
	\thanks{\copyright 2020 IEEE.  Personal use of this material is permitted.  Permission from IEEE must be obtained for all other uses, in any current or future media, including reprinting/republishing this material for advertising or promotional purposes, creating new collective works, for resale or redistribution to servers or lists, or reuse of any copyrighted component of this work in other works.}
}

\begin{document}

\bibliographystyle{ieeetr}

\maketitle

\begin{abstract}

The use of benchmarks is a widespread and scientifically meaningful practice to validate performance of different approaches to the same task. In the context of robot grasping the use of common object sets has emerged in recent years, however no dominant protocols and metrics to test grasping pipelines have taken root yet. In this paper, we present version 1.0 of GRASPA, a benchmark to test effectiveness of grasping pipelines on physical robot setups. This approach tackles the complexity of such pipelines by proposing different metrics that account for the features and limits of the test platform. As an example application, we deploy GRASPA on the iCub humanoid robot and use it to benchmark our grasping pipeline. As closing remarks, we discuss how the GRASPA indicators we obtained as outcome can provide insight into how different steps of the pipeline affect the overall grasping performance.

\end{abstract}



\section{INTRODUCTION}


\IEEEPARstart{I}{n} recent years, many robotic grasping pipelines have been proposed in the literature featuring consistent differences in hypotheses, methodology and experimental evaluation, in particular with respect to the objects and robotic platform used~\cite{bohg_data-driven_2014}.
Given such variability, reproducible test conditions, a standardized set of objects, a benchmarking protocol and a suite of metrics are fundamental to make fair performance comparisons.
Although a subset of the manipulation research community has already converged on a standard set of objects (i.e. the YCB object and model set~\cite{calli_ycb_2015}), a widespread protocol and a system of metrics for properly comparing different pipelines are still missing.  

Validation of candidate grasps in simulation alone with force closure quality measures~\cite{roa2015grasp} has been proven to be unreliable~\cite{kim2013physically}. Such a limitation, together with the lack of a dominant metric, led to the common practice of empirically testing grasp pipelines with a simple success rate over a given number of trials and objects~\cite{levine2018learning, mahler2017dex}. However, this kind of binary metric is somewhat limited, since it has no means of decoupling limitations of the algorithm itself from those of the test platform.

 \begin{figure}[t!]
	\centering
	\includegraphics[height=0.5\columnwidth]{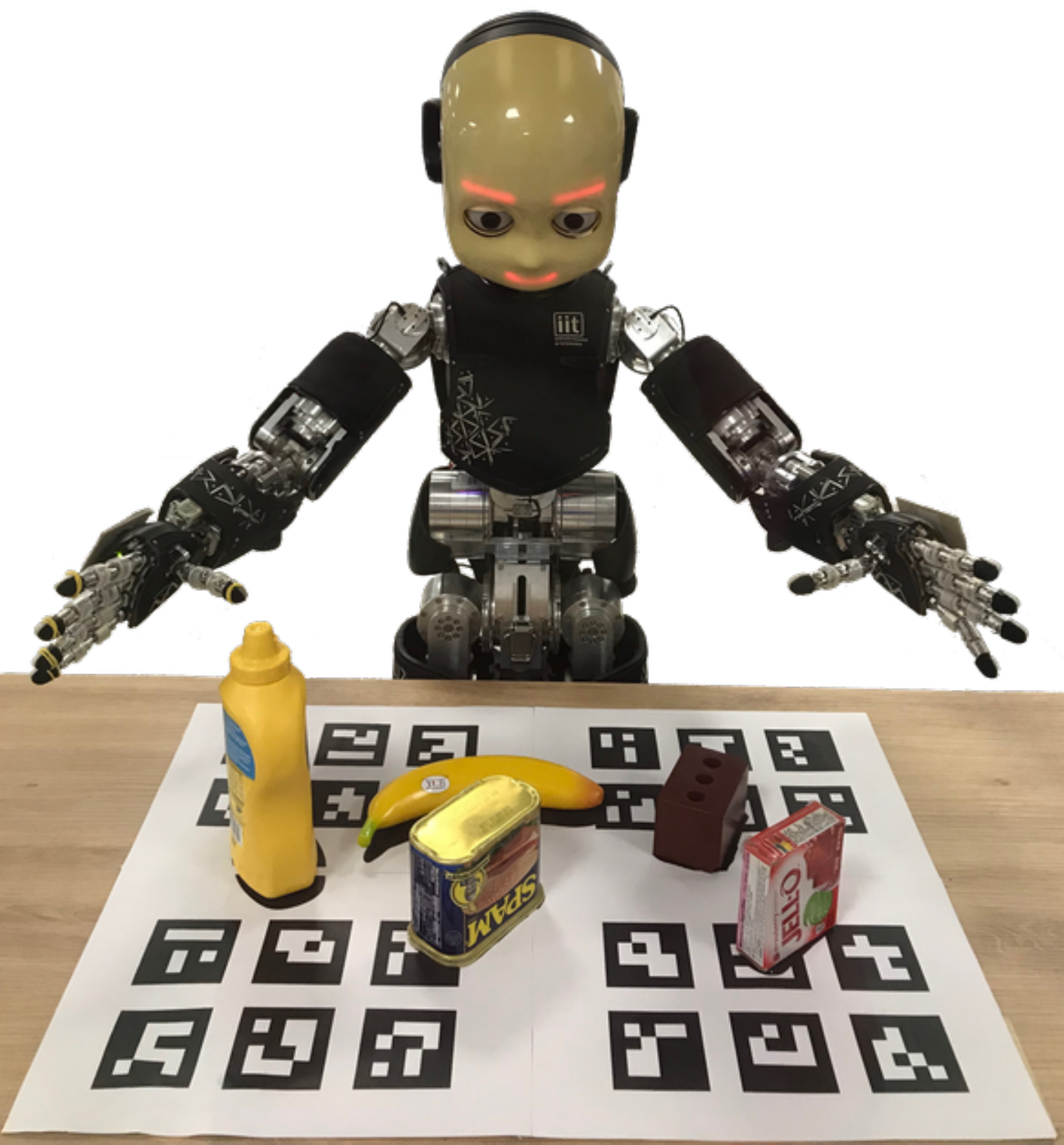}
	\caption{An example of the benchmark setting deployed on the iCub humanoid robot.\label{fig:setup}}
\end{figure}

In this paper we propose \textbf{GRASPA 1.0 (GRASPA is a Robot Arm graSping Performance benchmArk)}, a benchmarking protocol and a set of metrics to evaluate the performance of grasping pipelines. It aims to fairly compare methodologies tested on different robots by measuring and accounting for platform limitations that might hinder the overall performance. The proposed benchmark features:
 \begin{itemize}
 	\item Printable layouts of predefined grasping scenarios (populated with YCB object subsets) equipped with localization markers to enhance test reproducibility.
 	\item A protocol to assess the robot reachability and the calibration of the vision system within the defined grasping setup area.
	\item A widely-used grasp quality metric to evaluate candidate grasping poses before their physical execution.
 	\item A score to assess grasp stability during the physical execution on the robot.
	\item Possibility to benchmark the pipeline either in isolation or in clutter, with the definition of a further metric to evaluate the obstacle avoidance in the latter case.
 	\item A composite score to quantify the overall performance of the pipeline.
 \end{itemize}

We published on GitHub\footnote{\href{https://github.com/robotology/GRASPA-benchmark.}{https://github.com/robotology/GRASPA-benchmark}} the code for computing the benchmark scores and instructions on how to collect the required data.
Additionally, we made available a Docker container to ease installation and a cloud hosted environment to test the code without requiring any installation.

We employed GRASPA to assess the performance of the grasping pipeline proposed in~\cite{nguyen2018merging} using the iCub humanoid robot. The code we used to collect the data on the iCub is also available\footnote{\href{https://github.com/robotology-playground/GRASPA-test}{https://github.com/robotology-playground/GRASPA-test}} and can be used as an example procedure to collect the required data.

The paper is organized as follows. Section \ref{related-work} reviews relevant work concerning benchmarks available for grasping applications, including object sets and metrics. In Section \ref{benchmark} we outline the proposed benchmark. In Section \ref{application}, we provide an example by using GRASPA to benchmark a grasping pipeline on the iCub robot. Section \ref{conclusion} concludes the paper with some closing remarks and perspectives for future extensions of the benchmark. As part of this work, we attach to the submission a benchmark and a protocol document compiled according to the YCB benchmark templates\footnote{\label{note1}\href{http://www.ycbbenchmarks.com/protocols-and-benchmarks/}{http://www.ycbbenchmarks.com/protocols-and-benchmarks}}.

\section{RELATED WORK}
\label{related-work}

In recent years, the success of data-driven methods has brought new ideas and advancements in the field of robotic manipulation~\cite{bohg_data-driven_2014,mahler2017dex,kopickioneshot2016,levine2018learning,ten2017grasp}, at the same time pushing the community towards testing applications on common sets of both real objects~\cite{matheus_benchmarking_2010,kasper2012kit,calli_ycb_2015} and meshes~\cite{singh_bigbird_2014,ChangFGHHLSSSSX15} to develop benchmarking protocols. Despite the complexity of grasping pipelines and the variability in test setup design, however, most of the available benchmarks meant to be deployed on real robots are based on simple success/failure binary evaluation metrics$^{\ref{note1}}$.

Challenges such as the Amazon Picking Challenge~\cite{bell2015apc} and RoboCup@Home~\cite{stuckler_robocuphome_2012} proved to be quite effective in benchmarking entire autonomous pipelines by defining strict rules and tasks. However, in these contexts the tasks themselves are often difficult to reproduce and the number of accepted teams is typically small. 

The VisGraB benchmark~\cite{kootstra_visgrab_2012} presents a toolbox to evaluate vision-based grasp planners in simulation. VisGraB provides real stereo images of objects in various conditions and a software environment to analyze the quality of user-planned grasps in a simulated environment. However, it does not account for any real execution of the task, nor the type and performance of the manipulator and end effector. 

The ACRV benchmark~\cite{leitner_acrv} and the one published by Triantafyllou et al.~\cite{grocery_benchmark} tackle the issue of reproducibility by proposing a set of objects and  layouts for industrial shelving and pick and place applications. Both argue that physical execution of the task is essential in evaluating the performance of pick and place pipelines, although their protocols do not account for test platform limitations and the score metrics do not provide insight on the performance of single pipeline steps.

\section{BENCHMARKING PROTOCOL}
\label{benchmark}

In this Section we outline the proposed benchmarking protocol, focusing on the design of the grasping layouts, and the metrics to evaluate the individual pipeline steps.

\subsection{Benchmark Layouts}

\begin{figure}[t!]

	\subfigure[Benchmark Layout 0]{
		\label{fig:layouts-0}
	\includegraphics[width=0.44\columnwidth]{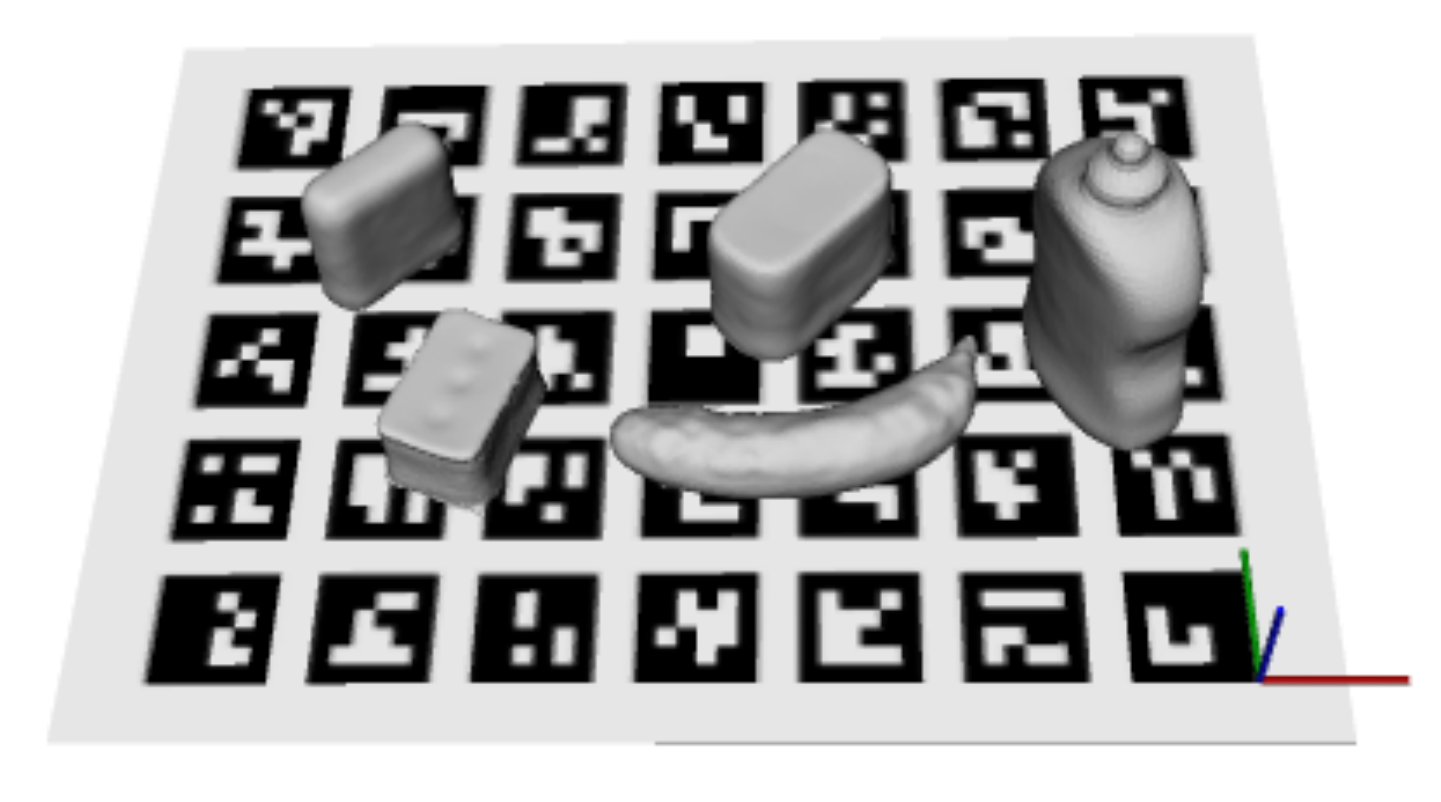}} \subfigure[Benchmark Layout 1]{
	\label{fig:layouts-1}
	\includegraphics[width=0.44\columnwidth]{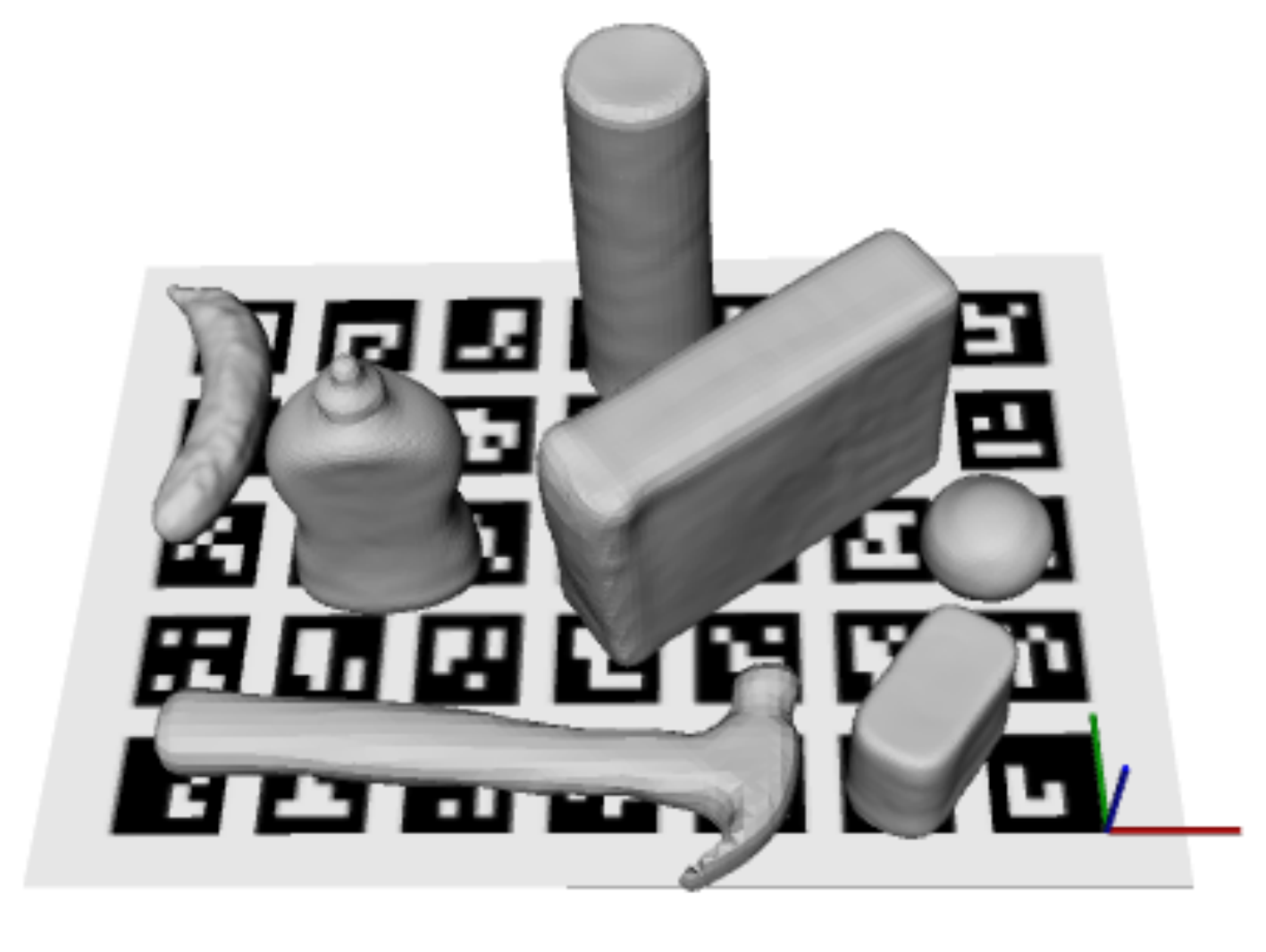}}

\subfigure[Benchmark Layout 2]{
	\label{fig:layouts-2}
	\includegraphics[width=0.44\columnwidth]{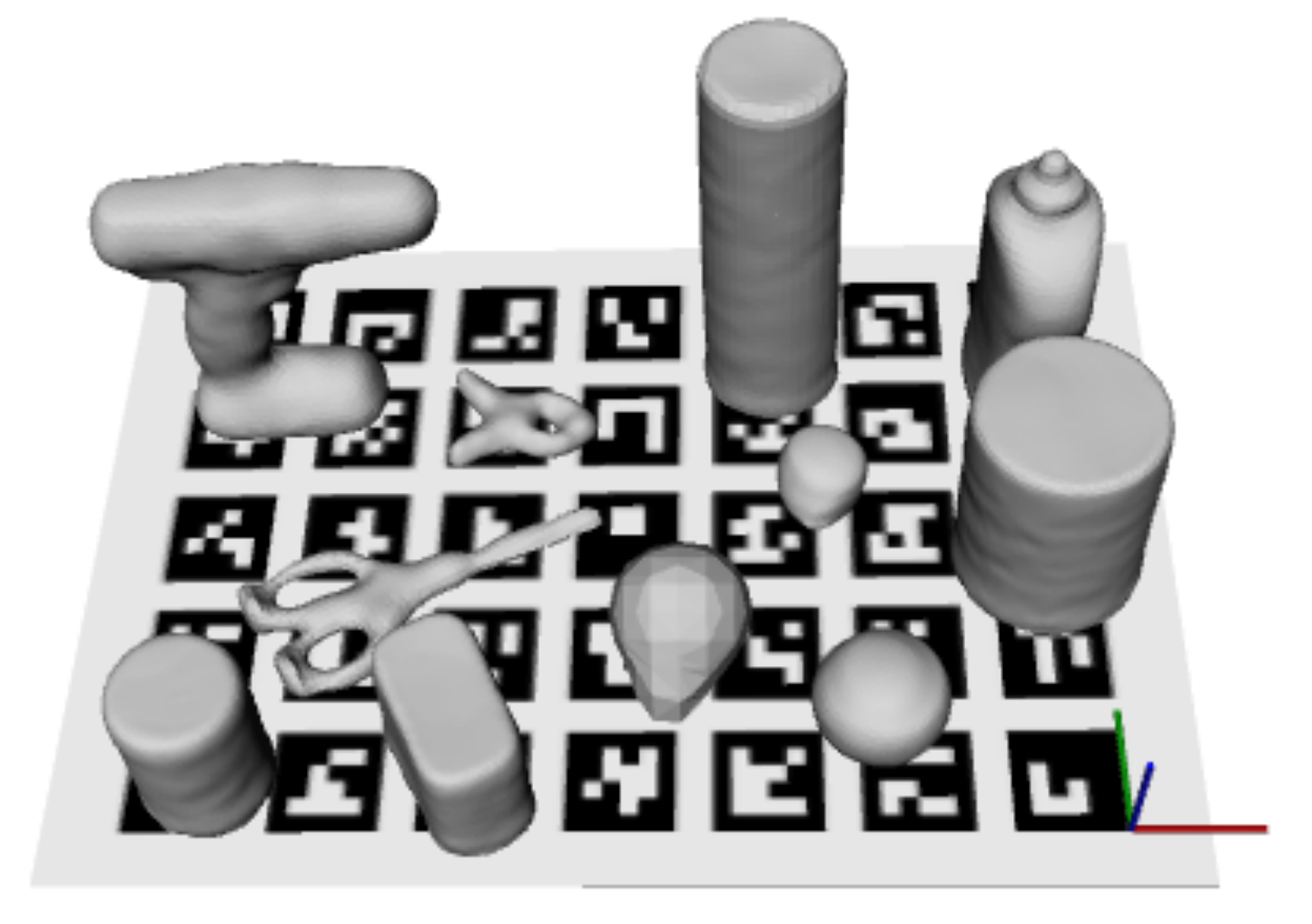}}\subfigure[Printable Layout 0]{
	\label{fig:printable-0}
	\includegraphics[width=0.44\columnwidth]{figures/layout_0_printout.pdf}}

	\caption{From (a) to (c): the 3D renders of the three layouts defined within the benchmark. (d) shows one of the provided printable boards that allow for reproducibile object placement on a physical setup. \label{fig:layouts}}
\end{figure}

GRASPA is designed to evaluate grasping pipelines on an area located in front of the robot with dimensions 594x420 mm (A2 standard paper size), resulting in the setup shown in Fig. \ref{fig:setup}. GRASPA uses a subset of the YCB object set (see Fig. \ref{fig:layouts}), selected in order to include a range of shapes, dimensions and challenges for the grasping task. We propose 3 scenarios of increasing complexity in terms of number, shape and pose of the included objects (see Fig. \ref{fig:layouts-0}, \ref{fig:layouts-1}, \ref{fig:layouts-2}). Moreover, GRASPA can evaluate pipelines that work both \textit{in isolation} (i.e. one object at a time in the layout) and \textit{in clutter\footnote{In this work, we refer to clutter as a situation where the objects are visually occluded (as long as a top down view is not used) and the presence of objects limits the task space of the robot while planning for grasp and avoiding collisions}} (i.e. all objects at the same time).  In the latter case, the added challenge is accounted for in the final score.

The 6D object poses are expressed with respect to the layout reference frame shown in Fig. \ref{fig:layouts-0}, \ref{fig:layouts-1}, \ref{fig:layouts-2}. To this end, an ArUco marker board~\cite{garrido2014automatic} is embedded in the printable layouts to enable the experimenter\footnote{From this point onwards, we refer to the experimenter as "the user"} to estimate the layout reference frame pose in a robust way. Users need to express all the information collected during the benchmark procedure with respect to the layout reference frame so as to be independent from the position of the physical board. Finally, we provide printable layouts of dimensions 594x420 mm  (i.e. A2 format) that include markers and object footprints (e.g. Fig. \ref{fig:printable-0}).

\subsection{Reachability within the Layout}
\label{reaching}

\begin{figure}[t!]

	\centering
	\includegraphics[width=0.35\columnwidth]{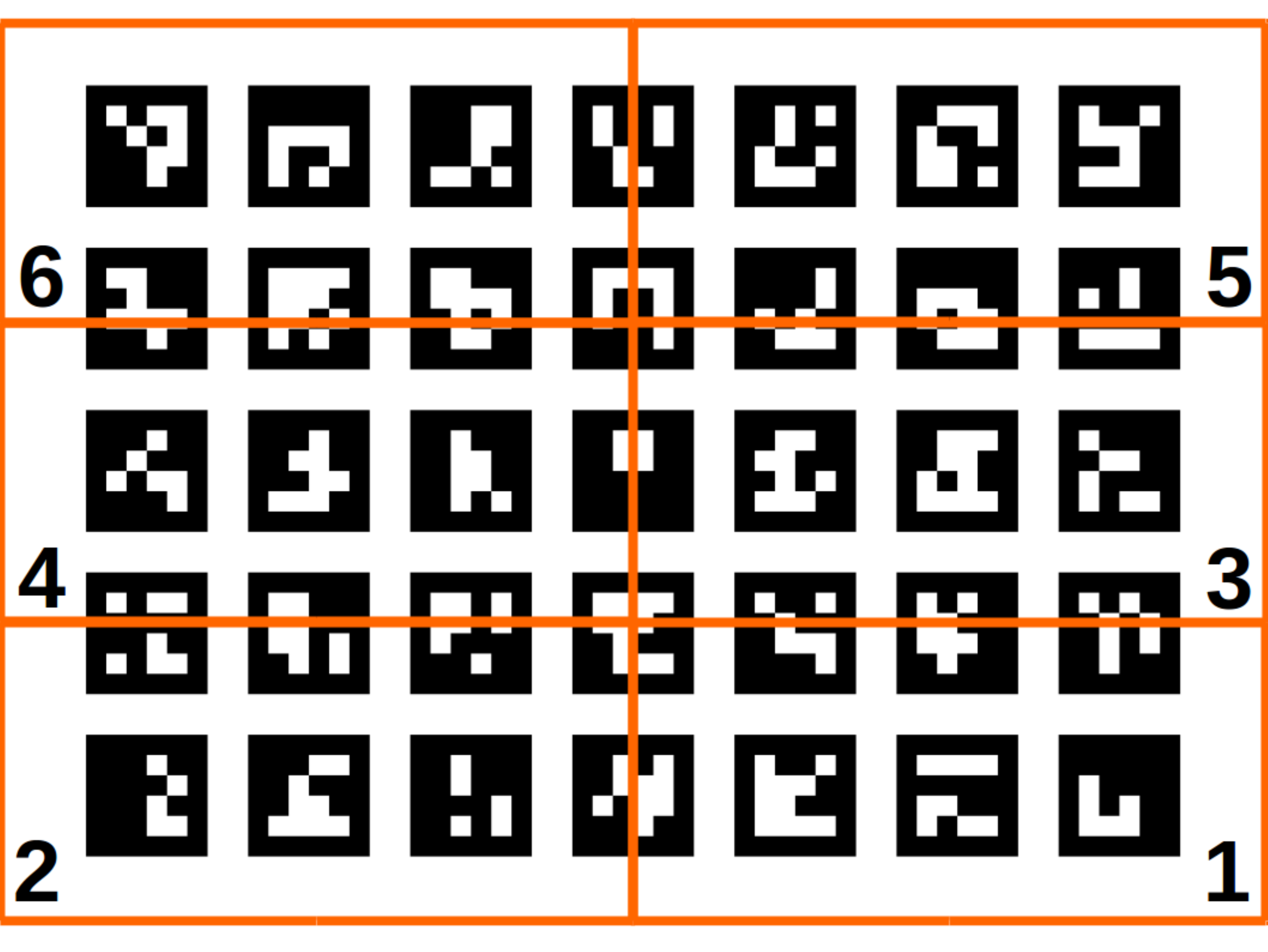}
	\caption{Regions used to determine the robot reachability and the calibration of the vision system within the layout.\label{fig:regions}}
\end{figure}

\begin{figure}
	\subfigure[Set no. 0 ]{
		\label{fig:reach-0}
		\includegraphics[width=0.3\columnwidth]{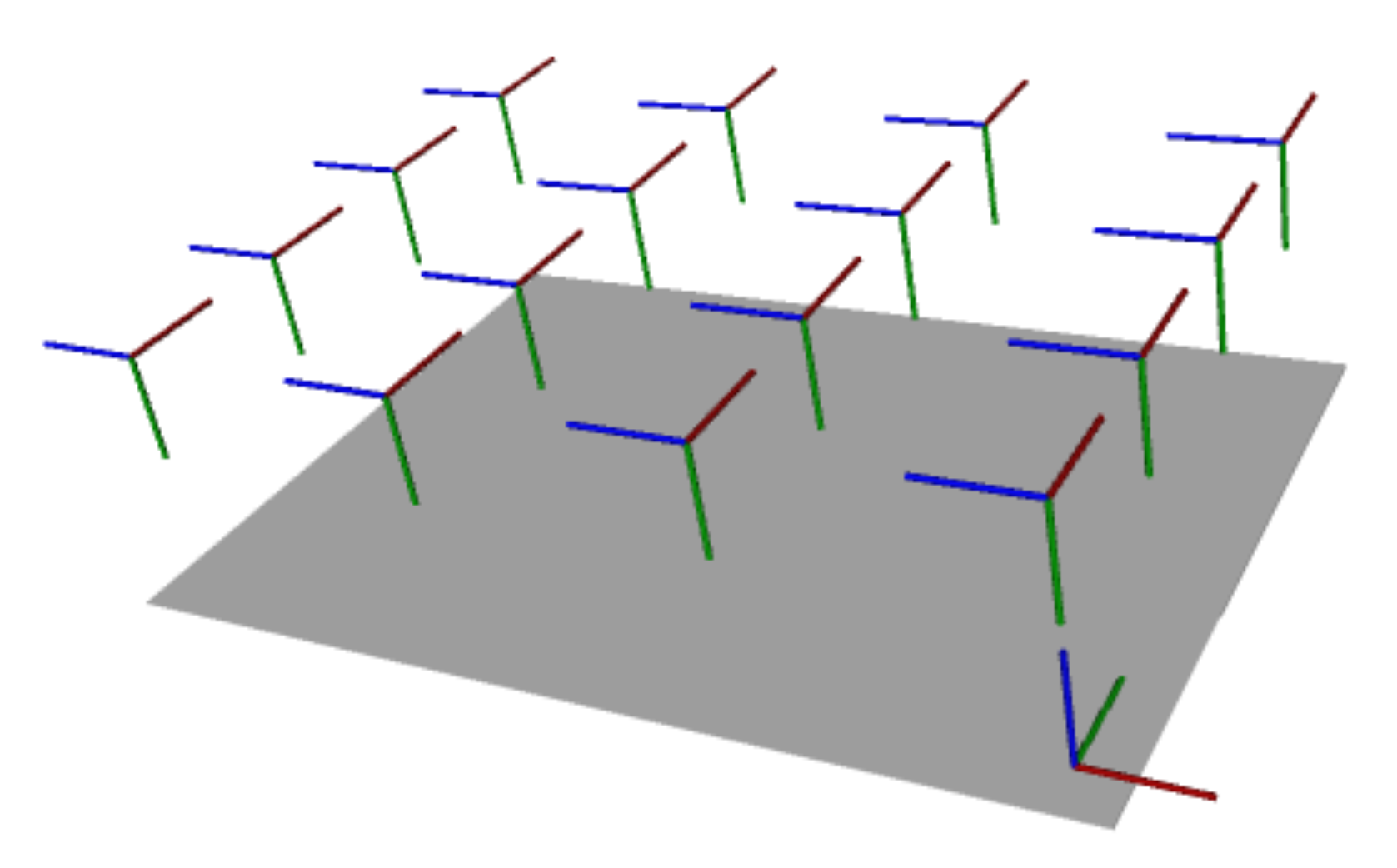}}
	\subfigure[Set no. 1]{
		\label{fig:reach-1}
		\includegraphics[width=0.3\columnwidth]{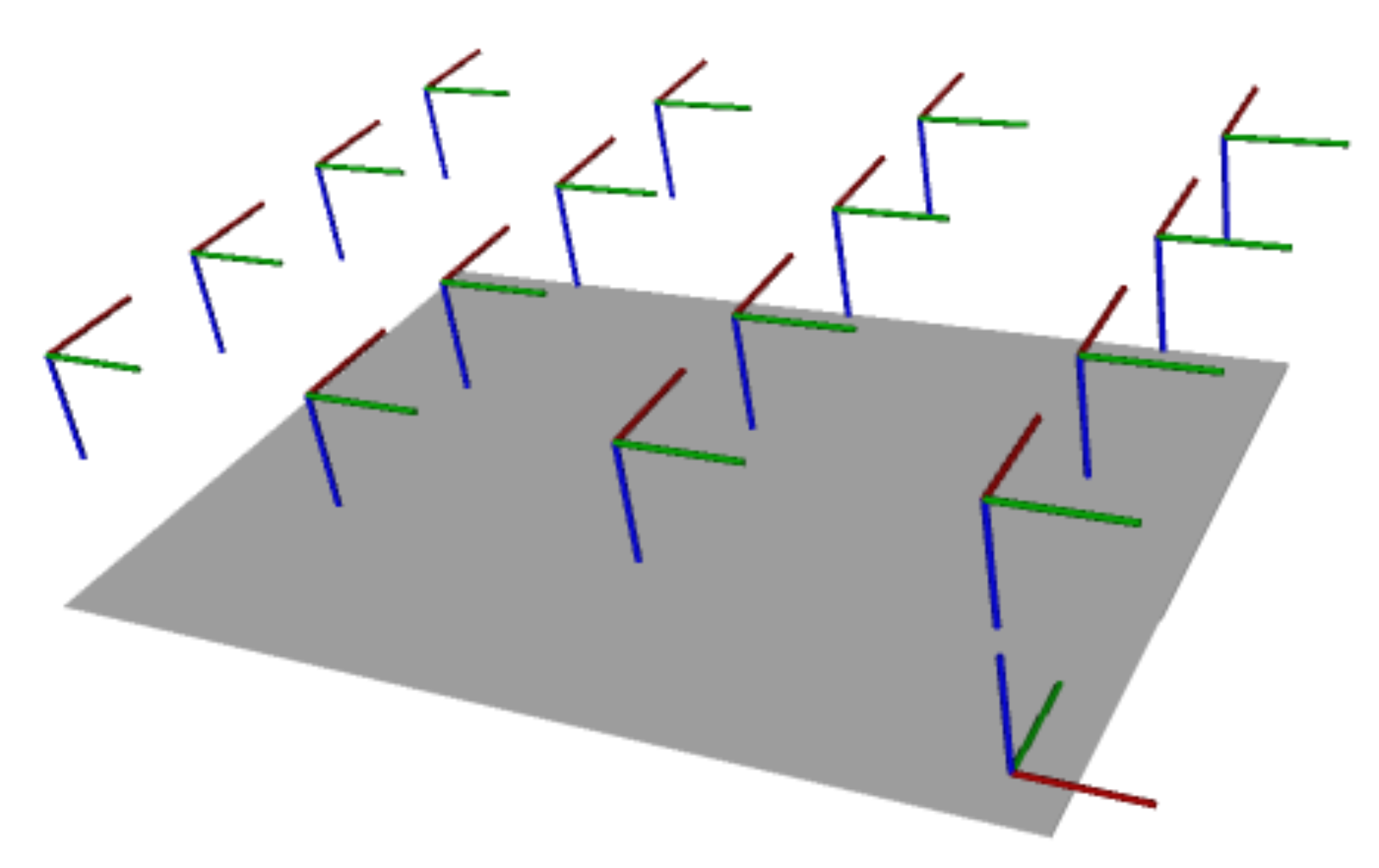}}
	\subfigure[Set no. 2]{
		\label{fig:reach-2}
		\includegraphics[width=0.3\columnwidth]{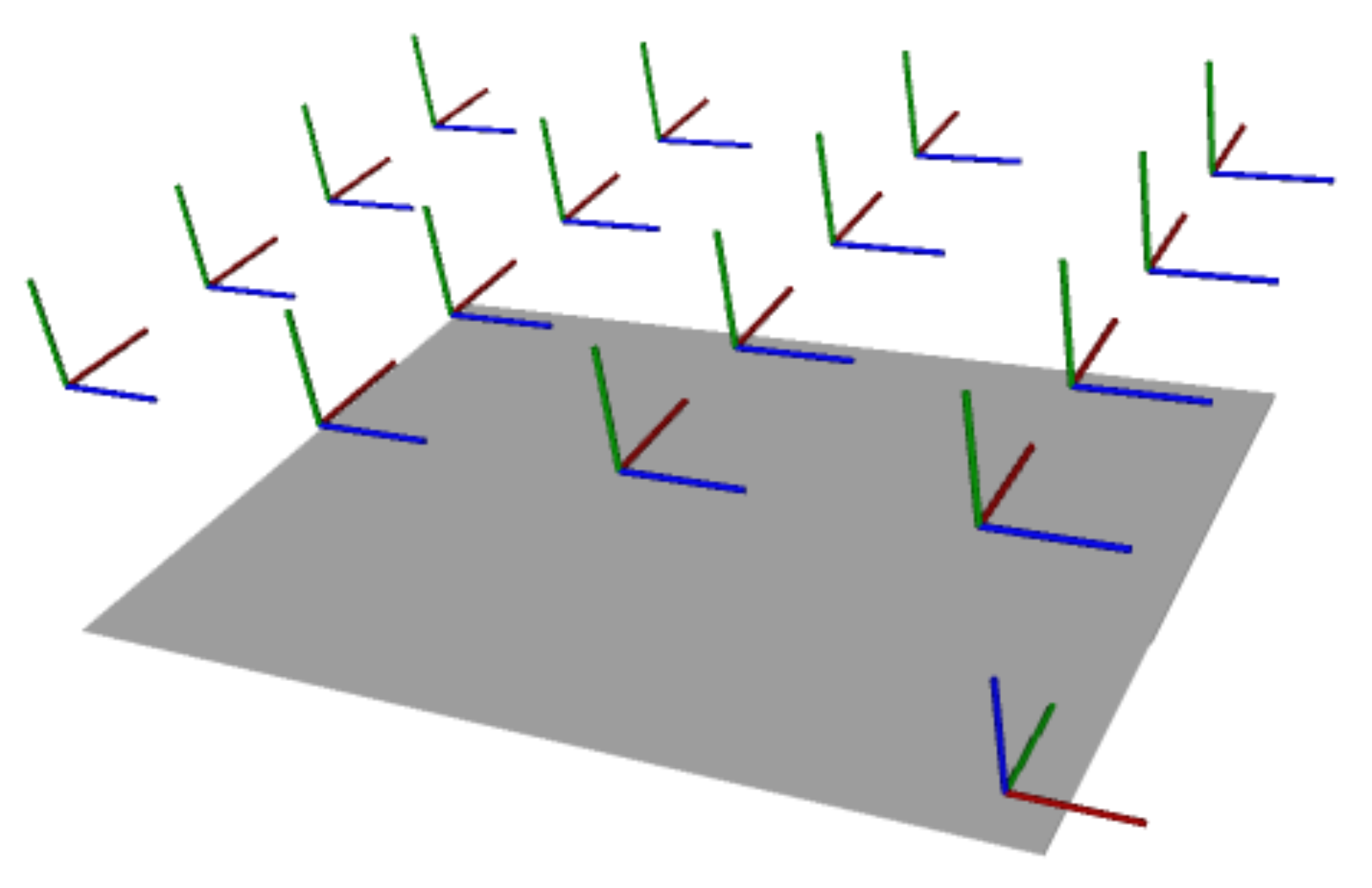}}

	\caption{Poses  defined for evaluating the robot reachability within the layout. Set of poses no. 1 (Fig. \ref{fig:reach-1}) is also used for testing the calibration of the vision system. \label{fig:reaching-poses}}
\end{figure}

Depending on the testing platform, the robot arm size, mechanical structure or joint range limits may impair the capability of the end-effector to reach some layout regions with accuracy. Therefore, grasps in these regions might fail regardless of the performance of the planner. To avoid penalizing planners for the limits of the test platform, an index of reachability over the layout area must be included in the benchmark. In GRASPA, we adopt an empirical approximation of such measure by dividing the layout area in 6 regions, each with a reachability score $S0_i$ for $i=1, \dots, 6$ (Fig. \ref{fig:regions}). The reachability score $S0_i$ for each region is defined over a set of poses uniformly distributed over the layout area with different orientations (Fig. \ref{fig:reaching-poses}). The user makes the robot reach (or attempt to) for these pre-defined poses and then acquire the ones actually reached by querying the forward kinematics. Poses placed on the boundary of contiguous regions are considered to belong to both regions.

The score $S0_i$ for the $i$-th region is given by: 
\begin{equation}
S0_i = \frac{N_i^{reached}}{N_i^{tot}} \in [0, 1],
\end{equation}
where $N_i^{reached}$ is the number of poses in region $i$ actually reached by the robot with a given accuracy and $N_i^{tot}$ is the number of poses belonging to the region $i$.

A pose $l$ is considered to be reached if the position  and orientation errors $(E^{p}_l$, $E^{o}_l)$  are smaller than the thresholds $(\tau_{p}^{r}, \tau_{o}^{r})$ defined by the user. In Section \ref{final_score} we elaborate more on such thresholds. The errors are computed as follows: 
\begin{align}
\label{error_s0}
E^{p}_l &= \Vert p^{reached}_l - p^{desired}_l\Vert,\\
E^{o}_l &= \sin{(\alpha^{error}_l)},
\end{align}
where $\alpha^{error}_l$ is the angle of the equivalent axis-angle representation of the matrix:
\begin{equation}
\label{r_s0}
R^{error}_l = R^{desired}_l \: R^{reached}_l,
\end{equation}
with $R^{desired}_l$ and $R^{reached}_l$ respectively the desired and reached orientation matrix relative to pose $l$~\cite{sciavicco2012modelling}.

For each benchmark layout $L \in \{0, 1, 2\}$, we associate to each object $k=1, \dots, N_{L}^{obj}$ (with  $N^{object}_L$ being the number of objects included in layout $L$) the reachability score $S0_{i^*}$ of the region $i^*$ where the object is located. For simplicity, an object belongs to the region its center of mass falls into. Thus, for each object $k=1, \dots, N_{L}^{objects}$ in each layout we obtain the \textit{reachability score $S0_k^L$}:

\begin{equation}
S0_k^L = S0_{i^*} = \frac{N_{i^*}^{reached}}{N_{i^*}^{tot}} \in [0, 1].
\end{equation}

\subsection{Camera Calibration within the Layout}
\label{camera}
Drawing a parallel to the reachability problem, GRASPA aims to assess the precision of the manipulator when reaching for poses acquired by the visual system in the camera reference frame.
Hence, our benchmark defines a camera calibration score $S1_i$ for each $i$-th region introduced in the Section \ref{reaching}. In order to evaluate the scores $S1_i$, the robot is asked to reach a subset of the poses defined for the reachability evaluation (Fig. \ref{fig:reach-1}). The 6D pose reached by the end-effector should be acquired through the vision system (e.g. by affixing a marker to the end-effector, in a known position and orientation with respect to the kinematic chain).

The score $S1_i$ is then computed as:
\begin{equation}
S1_i = \frac{N_i^{reached}}{N_i^{tot}} \in [0, 1],
\end{equation}
where $N_i^{reached}$ is the number of poses in region $i$ actually reached by the robot with a given accuracy and $N_i^{tot}$ is the number of poses belonging to the region $i$.

A pose $l$ is considered to be reached if the position  and orientation errors $(E^{p}_l$, $E^{o}_l)$ (computed according to Eq. \eqref{error_s0} - \eqref{r_s0}) are smaller than respective thresholds $(\tau_{p}^{c}, \tau_{o}^{c})$ defined by the user. The only difference with respect to the scores $S0_i$ is that the poses actually reached by the robot are acquired through the robot vision system and not from the forward kinematics.

Also in this case, for each benchmark layout $L \in\{0, 1, 2\}$, we associate to each object $k=1, \dots, N_{L}^{obj}$  the camera-calibration score $S1_{i^*}$ of the region $i^*$ where the object is located. Thus, for each object $k=1, \dots, N_{L}^{obj}$ we obtain the \textit{camera calibration score $S1_k^L$}:

\begin{equation}
S1_k^L = S1_{i^*} = \frac{N_{i^*}^{reached}}{N_{i^*}^{tot}} \in [0, 1].
\end{equation}

Since GRASPA layouts are defined with respect to the board reference frame, the benchmark protocol can be applied to grasping pipelines that do not process visual input (provided the user can reliably define a transform between the robot and the board reference frames). In such case, the benchmark does not take into account the scores $ S1_k^L$.

\subsection{Graspability}
\label{graspability}
Different robots might have diverse grasping capabilities due to the arm maximum payload and the end-effector design and size. Grasping pipelines should not be benchmarked on objects the robot cannot grasp or lift because of hardware limitations. GRASPA encodes this information in the \textit{graspability score} $S2_k^L = p_k^L \wedge g_k^L$, defined for each object $k=1, \dots, N^{object}_L$ in layout $L$. $p_k^L$ is 1 if the weight of the object is compatible with the robot payload and 0 otherwise. $g_k^L$ is 1 if the end effector aperture is larger than the smaller dimension of the object and 0 otherwise. For simple objects such as a box, this dimension is the shorter edge of the enclosing 3D bounding box, while for complex objects (e.g. the power drill) this can be the diameter of the grip. Objects can also be declared un-graspable by other criteria, if sufficient motivation is given.

\subsection{Grasp Quality}
\label{grasp_quality}

This index evaluates grasps planned by the pipeline before execution, regardless of reachability. GRASPA uses a metric that relies on computation of the Grasp Wrench Space (GWS) and Object Wrench Space (OWS)~\cite{borst_grasp_2004}. This metric, while not being the most robust to uncertainty~\cite{kim2013physically}, is still widely used in many grasping toolboxes such as Simox, OpenRAVE and GraspIt!~\cite{simox,diankov2008openrave,graspit}.

The user is required to provide the kinematic structure and the collision mesh model of their end-effector. Grasps have to be parametrized in terms of end effector pose and pregrasp configuration of the joints, making GRASPA compatible with both grippers and multifingered hands. Grasps are tested by first moving the end effector model to the desired pose with the desired pregrasp configuration, and then simulating the finger closure motion (in case of multifingered hands, joints are moved with equal velocity). When contact points are detected (via collisions between the object and end effector meshes), joints attached to the links that have collided are stopped. While this approach is straightforward for power grasps, pipelines that plan the contact locations need to be tested by setting the final hand configuration as a pregrasp.

We assume a \textit{hard point contact with friction} model with a fixed friction coefficient. Non-graspable objects (according to Subsection~\ref{graspability}) do not receive any score. The \textit{grasp quality} $\bar{S3}_k^L$ for each graspable object $k=1, \dots, N^{object}_L$  in layout $L$ can be expressed as

\begin{equation}
\bar{S3}_k^L = \frac{1}{T} \sum_{t=1}^{T}\Big( \frac {\bar{r} (GWS_{k, t})}{r (OWS_{k})}  \Big) \in [0, 1],
\end{equation}
where $\{\bar{r}(GWS_{k,t}) , r(OWS_k)\}$ are the radii of the largest spheres contained, respectively:

\begin{itemize}
	\item in the GWS defined by the $t$-th grasp planned for the $k$-th object. $\bar{r}(GWS_{t,k})$ is obtained by perturbing the grasping pose (before closing the fingers) in both position and orientation to ensure robustness, and then averaging the results;
	\item in the OWS of the $k$-th object, and is computed regardless of the grasp.
\end{itemize}

GRASPA v1.0 uses the implementation of the aforementioned metric included in GraspStudio~\cite{simox}.

\subsection{Grasp Execution and Stability}

GRASPA combines all the previously defined scores with grasp executions on physical robots. A \textit{binary success score} $\bar{S4}_k^L$ for each object $k=1, \dots, N^{object}_L$  in layout $L$ is evaluated over $T=5$ grasp executions:
\begin{equation}
\bar{S4}_k^L = \frac{1}{T}\sum_{t=1}^{T}\Big(S4_{k, t}^L\Big),
\end{equation}
where
\begin{equation*}
S4_{k, t}^L = \begin{cases} 1, & \mbox{if object $k$ has been grasped at trial $t$,}\\
0, & \mbox{otherwise.}
\end{cases}
\end{equation*}

The object is considered grasped if it can be lifted by $\delta_p=0.15$ m and held without falling for at least five seconds. Contact slip is acceptable as long as it does not ultimately cause the object to fall.
The score $\bar{S4}_k^L$ can be evaluated by executing both grasping in isolation or in the cluttered scene, assuming the same modality is kept for each object and layout.

Finally, the benchmark evaluates the stability of the grasp during the execution of a fixed trajectory. This trajectory simply consists of rotations around the end effector approach axis and in the vertical plane such axis passes through. Given the grasping pose $(p_{gr}, R_{gr})$, with $p_{gr} \in \mathbb{R}^3$ as position  and $R_{gr} \in SO(3)$ as the rotation matrix representing the orientation, the trajectory consists of 5 waypoints:	
\begin{align}
	  p_0 &= p_{gr} + \delta_p &  R_0 &= R_{gr} \\
	w_1:  p_1 &= p_0               &  R_1 &= R_{gr} \cdot R^+\\
	w_2:  p_2 &= p_0               &  R_2 &= R_{gr}\\
	w_3:  p_3 &= p_0               &  R_3 &= R_{gr} \cdot R^- \\
	w_4:  p_4 &= p_0               &  R_4 &= R_{gr}\\
	w_5:  p_5 &= p_0               &  R_5 &= R_{gr} \cdot R^\perp 
\end{align}
where $R^{+/-}$ represents a rotation of $\pm 45$ degrees around the approach axis of the end effector, and $R^\perp$ a rotation of 30 degrees (towards the table surface) in the vertical plane that contains this axis. The reference duration for each rotation trajectory is two seconds.
 We define the \textit{grasp stability} score $\bar{S5}_k^L$ for each object $k=1, \dots, N^{object}_L$ in layout $L$ over $T$ as:
\begin{equation}
\bar{S5}_k^L = \frac{1}{T}\sum_{t=1}^{T}\Bigg(\frac{N_{w, t}^{reached}}{N_{w}^{tot}}\Bigg) \in [0, 1],
\end{equation}
where $N_{w, t}^{reached}$ is the number of the trajectory waypoints reached without dropping the object at trial $t$ and $N_{w}^{tot} = 5$ is the total number of the trajectory waypoints. Again, contact slip is acceptable if it does not lead to a fall.

If the pipeline under test allows for it, GRASPA can measure its ability to grasp while avoiding other objects. We define the \textit{obstacle avoidance} score $\bar{S6}_k^L$ for $k=1, \dots, N_L^{obj}$ over $T$ trials:
\begin{equation}
\bar{S6}_k^L = \frac{1}{T} \sum_{t=1}^{T} \Bigg( 1 - \frac{N_{hit, t}}{N_L^{obj}}\Bigg) \in [0, 1],
\end{equation}

where $N_{hit, t}$ is the number of objects hit by the robot while approaching the target object at trial $t$. The score is 1 if the robot is able to avoid all the objects and 0 if it collides with every object. If no obstacle avoidance is accounted for, tests must use single objects and S6 is not computed.

\section{Reporting benchmark scores}
\label{reporting_scores}

In this Section, we explain how the single step metrics are combined into a single composite score. We outline how the benchmark scores are reported, giving guidelines on how to interpret the outcome and how to choose the required user-defined thresholds.

\subsection{Final composite score and summary table}
\label{final_score}

All the scores proposed thus far contribute to the computation of a composite score $\bar{S}_L$ to evaluate the grasping pipeline performance in each layout $L$, accounting for the limits of the testing platform.
To this aim, the final score is computed considering only objects $m=1, \dots,  M_{L}^{obj}$ such that:
\begin{itemize}
	\item $m$ is graspable by the robot, i.e. $S2_{m}^L=1$;
	\item $m$ is in a reachable region, i.e. $S0_{m}^L> 0.5$. A region is not considered to be reachable if less than half the test poses were not reached with precision;
	\item $m$ is in a region with a good calibration of the vision system, where at least half the calibration poses were reached with acceptable precision, i.e. $S1_{m}^L> 0.5$.
\end{itemize}
 The expression of the final score $\bar{S}_L$ is the following:
\begin{equation}
\bar{S}_{L} = \frac{1}{M_L^{obj}}\sum_{m = 1}^{M_L^{obj}}{\bar{S}_{m}^L}  ,
\end{equation}
where, if benchmarking with objects in isolation:
\begin{equation*}
\bar{S}_m^L  = \frac{1}{T}\sum_{t=1}^{T}{(S3_{m, t}^L + S5_{m, t}^L) \cdot S4_{m, t}^L} \in [0, 2],
\end{equation*}
whereas, if benchmarking in clutter:
\begin{equation*}
\bar{S}_m^L  =
 \frac{1}{T} \sum_{t=1}^{T}{(S3_{m, t}^L + S5_{m, t}^L + S6_{m, t}^L) \cdot S4_{m, t}^L} \in [0, 3].
\end{equation*}

where $L$ indicates the layout, $m$ indicates the object and $t$ indicates the trial, $S3_{m, t}^L$ is the \textit{grasp quality} score, $S5_{m, t}^L$ is the \textit{grasp stability} score, $S6_{m, t}^L$ is the \textit{obstacle avoidance} score (if the pipeline allows for it), and $S4_{m, t}^L=1$ only if the object has been successfully grasped at trial $t$. The scores computed by the benchmark are summarized in Table \ref{tab:scores}.

\begin{table*}[t!]
	\centering
	\begin{tabular*}{1.01\linewidth}{lll}
		Score formula & Score name & Meaning\\
		\hline
		\begin{minipage}{3.5in}\vspace{3pt}
			$S0_k^L = S0_{i^*} = \frac{N_{i^*}^{reached}}{N_{i^*}^{tot}} \in [0, 1]$ \end{minipage}
		& Reachability score & \Bstrut \begin{minipage}{2.0in} \vspace{2pt}Accounts for whether the object is located in a region characterized by a good reachability of the robot. \end{minipage} \vspace{2pt} \\

		\hline
		\begin{minipage}{3.5in}\vspace{3pt}
			$S1_k^L = S1_{i^*} = \frac{N_{i^*}^{reached}}{N_{i^*}^{tot}} \in [0, 1]$ \end{minipage}
		& Camera-calibration score & \begin{minipage}{2.0in} \vspace{2pt}Accounts for whether the object is located in a region characterized by a good calibration of the vision system.\end{minipage} \vspace{2pt}\\

		\hline
		\begin{minipage}{3.5in}\vspace{3pt}
			$S2_k^L = p_k^L \wedge g_k^L \in \{0, 1\}$ \end{minipage}
		& Graspability score & \begin{minipage}{2.0in}\vspace{2pt}Accounts for whether the object can be physically grasped and lifted by the robot, considering its shape and weight.\end{minipage} \vspace{2pt}\\

			\hline
			\begin{minipage}{3.5in}\vspace{3pt}
				$\bar{S3}_k^L = \frac{1}{T} \sum_{t=1}^{T}\Big( \frac {\bar{r} (GWS_{k, t})}{r (OWS_{k})}  \Big) \in [0, 1]$\end{minipage}
			& Grasp quality score & \begin{minipage}{2.0in}\vspace{2pt}Accounts for how contacts are distributed on the object by simulating grasp closure in simulation and computing the grasp wrench space.\end{minipage}\vspace{2pt}\\

		\hline\
		\begin{minipage}{3.5in}\vspace{3pt}
			$\bar{S4}_k^L = \frac{1}{T}\sum_{t=1}^{T}\Big(S4_{k, t}^L\Big) \in [0, 1]$\end{minipage}
		& Binary success score & \begin{minipage}{2.0in}\vspace{2pt}Accounts for whether the robot actually managed to grasp the object in real tests.\end{minipage}\vspace{2pt}\\

		\hline
		\begin{minipage}{3.5in}\vspace{3pt}
			$\bar{S5}_k^L = \frac{1}{T}\sum_{t=1}^{T}\Bigg(\frac{N_{w, t}^{reached}}{N_{w}^{tot}}\Bigg) \in [0, 1]$\end{minipage}
		& Grasp stability score & \begin{minipage}{2.0in}\vspace{2pt}Evaluates the stability of the grasp during the execution of a fixed trajectory.\end{minipage}\vspace{2pt}\\

		\hline
		\begin{minipage}{3.5in}\vspace{3pt}
			$\bar{S6}_k^L = \frac{1}{T} \sum_{t=1}^{T} \Bigg( 1 - \frac{N_{hit, t}}{N_L^{obj}}\Bigg) \in [0, 1]$\end{minipage}
		& Obstacle avoidance score  & \begin{minipage}{2.0in}\vspace{2pt}(Only in cluttered mode) Accounts for how many objects the robot has hit while executing the grasp.\end{minipage}\vspace{2pt}\\

		\hline

		\begin{minipage}{3.5in}\vspace{3pt}
			$\bar{S}_{L} = \frac{1}{M_L^{obj}}\sum_{m = 1}^{M_L^{obj}}{\bar{S}_{m}^L}$ where\\
			in isolation: $	\bar{S}_m^L  = \frac{1}{T}\sum_{t=1}^{T}{(S3_{m, t}^L + S5_{m, t}^L) \cdot S4_{m, t}^L} \in [0, 2]$ \\
			in the clutter: $\bar{S}_m^L  = \frac{1}{T}\sum_{t=1}^{T}{(S3_{m, t}^L + S5_{m, t}^L + S6_{m, t}^L) \cdot S4_{m, t}^L} \in [0, 3]$ \end{minipage}
		& Final per object score & \begin{minipage}{2.0in}\vspace{2pt} Combines all the scores in order to evaluate the grasping pipeline performance taking into account any limitation of the robotic platform used in real world tests.  \end{minipage}\vspace{2pt}\\

		\hline
		\vspace{1pt}
	\end{tabular*}
	\caption{Summary of the benchmark scores.}
	\label{tab:scores}
\end{table*}

The final output of the benchmark consists of a summary Table (\ref{tab:benchmark-icub} for an example).  In the second column, the value of the final score $\bar{S}_L$ for each layout $L=0,1,2$ is reported. In the rest of the table, each row collects all the scores computed for each object $k = 1. \dots. N_L^{obj}$. Analyzing such scores can give insight about the performance of different parts of the grasping pipeline, down to the hardware. For instance, if the \textit{grasp quality score} $\bar{S3}_k^L$ is high but the robot could not grasp the object ($\bar{S4}_k^L=0$), the \textit{reachability score} $S0_k^L$ and the \textit{camera-calibration} score $S1_k^L$ can outline whether the vision system calibration or the robot reachability are to blame for the failure in the execution of the grasp. On the other hand, if $\bar{S3}_k^L$ is low, but $\bar{S4}_k^L$ and $\bar{S5}_k^L$ are large, this may indicate that the physical execution is able to compensate for the poor grasp quality (e.g. the gripper is compliant and can conform to the object, or the object pose changes during the grasp execution).

\subsection{Defining reachability and camera calibration thresholds}
\label{benchmark_thresholds}
As previously mentioned, GRASPA requires position and orientation thresholds used during the reaching test $((\tau_p^r, \tau_o^r)$, see Paragraph \ref{reaching}) and the camera calibration test ($(\tau_p^c, \tau_o^c)$, see Paragraph \ref{camera}). Since GRASPA is meant to adapt to different robot platforms, these thresholds cannot be fixed a priori by the benchmark and have to be chosen by the user according to the robot platform and vision system. 
$(\tau_p^r, \tau_o^r)$ define how precise the robot kinematics is over the GRASPA layout space. For dexterous and precise arms (e.g. industrial manipulators), small values of the reachability thresholds are advisable (e.g. $\tau_p^r = 0.005 \text{ m},  \tau_o^r = 0.1 \text{ rad}$). For less precise robots (e.g. research-oriented platforms such as iCub, PR2, Baxter) higher values are needed (e.g. $\tau_p^r = 0.02 \text{ m},  \tau_o^r = 0.5 \text{ rad}$). On the other hand, $(\tau_p^c, \tau_o^c)$ depend on camera resolution and the method used to visually infer the end effector pose. Upper bounds on these parameters are $\tau_p^r = 0.05 \text{ m},  \tau_o^r = 0.5 \text{ rad}$, that we found borderline acceptable for a 320x240 resolution camera.

Note that the aforementioned thresholds are mostly useful in the presence of hardware limits, inverse kinematics solver or calibration. In this scenario, low thresholds will likely mark some regions as unreachable or not well calibrated and will allow only grasps in regions where their execution can be more precise. With high thresholds, grasps will be executed and scored in regions where lack of precision might lead to unstable grasps and unfair scoring.

\section{EXAMPLE OF APPLICATION}
\label{application}
In this Section, we show an example application of the GRASPA protocol. We evaluated the grasping pipeline proposed in~\cite{nguyen2018merging} by using the iCub humanoid robot~\cite{icub} as the testing platform. We evaluated right-handed grasps performed \textit{in isolation}, although GRASPA is extendable to multi-armed planning approaches.

\subsection{Cardinal Point Grasps}
\label{cardinal_point_grasps}
Our grasping pipeline can be briefly summarized as follows.
\begin{itemize}
	\item \textbf{2D segmentation. }Using the monocular image stream coming from iCub, we adapt an off-the-shelf Tensorflow implementation~\cite{matterport_maskrcnn_2017} of Mask R-CNN~\cite{he2017mask} in order to obtain segmentation masks of the objects. We use a ResNet-50 backbone pre-trained on MS Coco, further training it on a subset of YCB-Video~\cite{xiang_posecnn_2017} and then fine-tuning it on a custom synthetic dataset. The latter was obtained by augmenting real images with YCB object crops following the Cut, Paste and Learn approach~\cite{dwibedi_cut_2017} enhanced with segmentation masks. The dataset features the 16 YCB objects used in GRASPA as classes, and ArUco marker crops as distractors.
	\item \textbf{Object modeling. }Partial object point clouds are obtained from segmentation masks through the robot stereo vision. As described in~\cite{nguyen2018merging}, we approximate the object with the smallest superquadric fitting the point cloud. The superquadric and its 6D pose are estimated by solving a constrained optimization problem, imposing one of the axes of the superquadric to be perpendicular to the table surface.
	\item \textbf{Grasp planning. } We generate grasping pose candidates from the cardinal points of the superquadric (i.e. where axes intersect the surface). The candidates are then ranked according to the superquadric and hand size, and the capability of the robot to reach them with sufficient accuracy~\cite{nguyen2018merging}.
\end{itemize}

\subsection{Data collection}
Hereafter, we briefly explain the procedure we followed to collect the data required by the benchmark from the physical robot. More information, together with a sample code as well as the reachability and calibration poses and object poses, is available online\footnote{\href{https://github.com/robotology-playground/GRASPA-test}{https://github.com/robotology-playground/GRASPA-test}}.
\subsubsection{Reachability score $S0$}
\begin{figure}
 	\centering
 	\subfigure[Desired poses (set no. 1) ]{
 		\label{fig:reach-0-noboard}
 		\includegraphics[width=0.4\columnwidth]{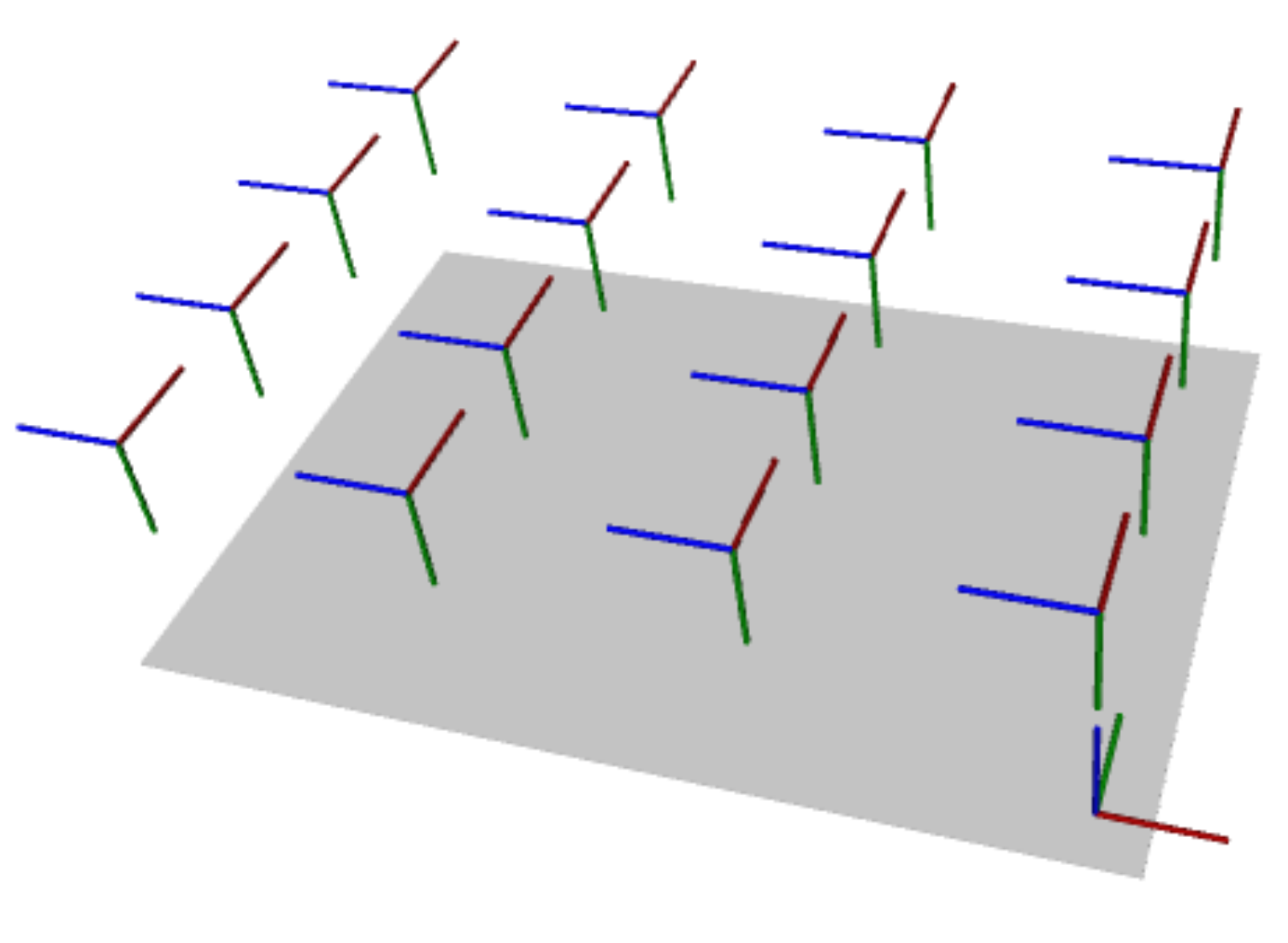}}
 	\subfigure[Reached poses (set no. 1)]{
 		\label{fig:reached-poses}
 		\includegraphics[width=0.4\columnwidth]{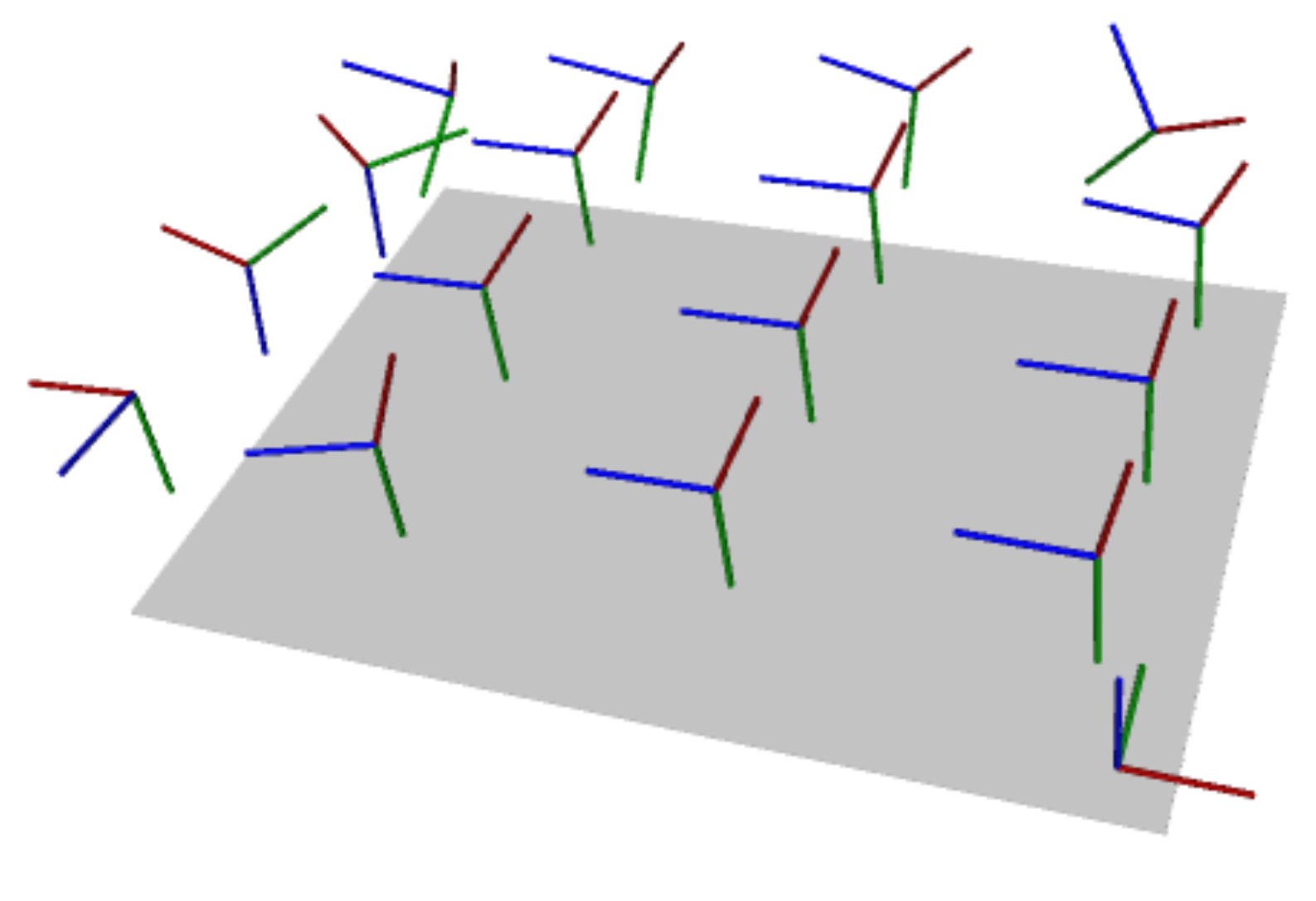}}

 	\caption{Reachability test results: comparison between the objective poses and those actually reached by iCub. \label{fig:icub-reach}}
 \end{figure}

Data for the computation of the reachability scores $S0_i$ has been acquired by having iCub reach  the poses defined within the benchmark with the right hand, querying the forward kinematics to obtain the poses actually reached. We used OpenCV to estimate the pose of the layout marker boards (Fig. \ref{fig:layouts}) with respect to the robot. We used this information to express the target poses in the robot reference frame and save the reached poses in the layout reference frame. Fig. \ref{fig:icub-reach} shows some samples of the outcome.

 \subsubsection{Camera-calibration score $S1$}
We followed the same procedure just outlined for the reachability score. Instead of acquiring the reached pose through the forward kinematics, we resorted to visual detection of two ArUco markers located on the back and the side of the hand.

\subsubsection{Graspability $S2$} We considered an object to be graspable by the iCub if at least one of its dimensions was smaller than the iCub hand aperture and its weight was compatible with the maximum arm payload (0.5 kg). We considered un-graspable by iCub objects that have a very low profile (i.e. scissors, clamp) when laid flat on the table.

\subsubsection{Grasp Quality $S3$}
\begin{figure}
	\centering
	\includegraphics[width=0.7\columnwidth]{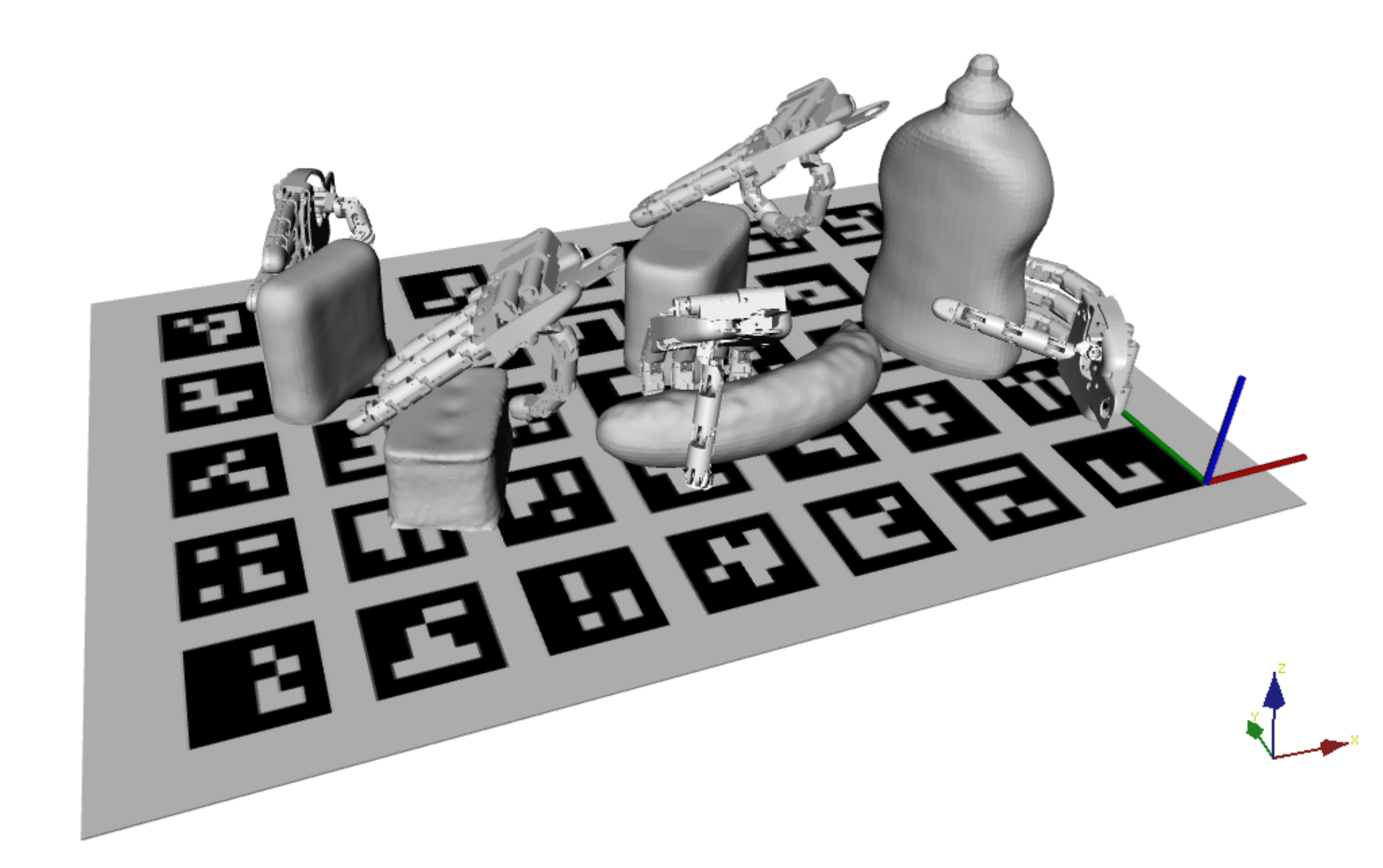}
	\caption{Rendering of the grasp poses planned for layout 0 with the tested algorithm. For visual clarity, only one pose is rendered for each object. \label{fig:gq_render_layout_0}}
\end{figure}

For each object visible in each layout, we planned for $T = 5$ 6D grasping poses according to Section~\ref{cardinal_point_grasps}, expressed in the layout reference frame by using the estimated pose of the ArUco marker board. We used the iCub hand model packaged with the GraspStudio suite~\cite{simox}. A graphical rendering of some of the planned poses can be seen in Figure~\ref{fig:gq_render_layout_0}.

\subsubsection{Binary Success and Stability scores ($S4, S5$)}
We executed in isolation the $T=5$ grasps computed by the algorithm for each object. Whenever the robot managed to grasp the object we also had it execute the trajectory defined in Section \ref{graspability}. We added a layer of rubber on the robot fingertip to actually have friction on the contact points. In these tests, we did not perform the last waypoint of the stability trajectory.

\subsection{Results and Discussion}
Table \ref{tab:benchmark-data} collects the user-defined parameters we used for data acquisition and score computation. The threshold values $(\tau_p^r, \tau_o^r, \tau_p^c, \tau_o^c)$ have been chosen considering the reachability and the visual calibration limits of the iCub. The pipeline under test plans for power grasps (i.e. it computes a pose for the hand palm and not for each fingertip) and is, therefore, able to deal with reachability errors in position  of $\tau_p^r = 0.02 \: m$ and in orientation of $\tau_o^r = 0.5 \: rad$. In order to deal with higher errors in the calibration of the vision system ($\tau_p^c = 0.045$ and $\tau_o^c = 0.8$), we made use of a calibration map obtained by kinesthetically teaching the robot the correction to be applied over a set of poses.

Computed scores are reported in Table \ref{tab:benchmark-icub}. We highlight:  1) the values of the reachability score $S0_k^L$ and the camera calibration score $S1_k^L$ when $S0_k^L < 0.5$ and $S1_k^L < 0.5$ (the object is in a region unreachable by the robot or with an unacceptable visual calibration error);  2) the value of the graspability score when the object is not graspable,
i.e. $\bar{S2}_k^L = 0$.
In these cases, the final score $\bar{S}_k^L$ is not computed by the benchmark and is replaced with the placeholder N/A. Because of the proximity of some objects to the robot torso, the stereo vision could not reliably acquire partial point clouds. In these cases, no further score is reported.

Table \ref{tab:benchmark-icub} shows how our benchmark properly evaluates the grasping pipeline without penalizing its performance wherever the test platform proved its limits. A meaningful example is the \textit{foam brick} in layout 0. The grasp quality score $\bar{S3}_k^L$ is good, meaning that the algorithm computes proper grasping poses for the object. However, in practice, the robot could grasp the object only once over the 5 trials ($\bar{S4}_k^L=0.2$). The reason of such failure can be attributed to the poor vision system calibration in the region of the object ($S1_k^L=0.25$). Therefore, the foam brick scores do not contribute to the computation of the final composite score.
On the other hand, other objects (e.g. potted meat can, cracker box and tennis ball) have low $\bar{S3}_k^L$ in layouts 0 and 1 but show higher values for $\bar{S4}_k^L$ and $\bar{S5}_k^L$. We observed this to be caused by the mechanical underactuation (not modeled in the GraspStudio~\cite{simox} environment) in the iCub hand that allows the fingers to conform to the object. 

\begin{table*}[t!]
	\centering
	\begin{tabular}{lllllll}	
Robot & End-effector & Modality & $\tau_{p}^{r}$ & $\tau_{o}^{r}$ & $\tau_{p}^{c}$ & $\tau_{o}^{c}$ \\
\hline\\
iCub & Right hand & In isolation & 0.02 m & 0.5 rad & 0.045 m & 0.8 rad \\
	\vspace{1pt}
	\end{tabular}
	\caption{User-defined parameters used during benchmarking procedure.}
	\label{tab:benchmark-data}
\end{table*}

\begin{table*}[t!]
	\centering
	\begin{tabular}{lc|l|c|c|c|c|c|c|r|}
		\multicolumn{2}{l}{Layout}  & \multicolumn{8}{l}{Per object scores}\\
		\hline\\

		Layout 0 & $\bar{S}_L$ &Object  & $S0_k^L$ & $S1_k^L$  & $S2_k^L$& $\bar{S3}_k^L$& $\bar{S4}_k^L$& $\bar{S5}_k^L$ & $\bar{S}_k^L$\\
		& 0.60 & banana & 1.0 & 0.75  & 1.0 & 0.32 & 0.8 &  0.2 & 0.36 \\
		& & foam brick &  0.75 & \textbf{0.25}  & 1.0 & 0.27 &0.2  & 0.2 & N/A \\
		& & gelatin box & 0.75  & \textbf{0.25}  & 1.0 & 0.07 &0.2  & 0.0 & N/A \\
		& &  mustard bottle &  1.0 & 1.0 & 1.0 & 0.15 & 1.0  & 0.8  & 0.95 \\
		& & potted meat can & 1.0 & 1.0  & 1.0 & 0.01 & 0.8  & 0.45 & 0.46 \\
		&&&&&&&&&\\

		Layout 1 & $\bar{S}_L$ &Object & $S0_k^L$ & $S1_k^L$  & $S2_k^L$& $\bar{S3}_k^L$& $\bar{S4}_k^L$& $\bar{S5}_k^L$ & $\bar{S}_k^L$\\
		& 0.70 &  banana & \textbf{0.25} & \textbf{0.0}  & 1.0 & 0.19 & 0.0 & 0.0 & N/A \\

		& & hammer & 0.75 & \textbf{0.25}  & \textbf{0.0} & N/A & N/A & N/A & N/A \\
		& & chips can & 0.5 & 0.5  & 1.0 & 0.25 & 1.0 & 1.0 & 1.25 \\

		& & tennis ball & 1.0 & 1.0  & 1.0 & 0.23 & 0.2 & 0.2 & 0.29 \\
		& & cracker box & 1.0 & 1.0 & 1.0 & 0.04 & 0.8 & 0.5 & 0.54 \\
		& & mustard bottle & 0.75 & \textbf{0.25}  & 1.0 &0.23 & 0.8 & 0.15 & N/A \\
		& & potted meat can & 1.0 & 0.75  & 1.0 & 0.01 & 0.8 & 0.7 & 0.71 \\

		&&&&&&&&&\\
		Layout 2 & $\bar{S}_L$ &Object & $S0_k^L$ & $S1_k^L$  & $S2_k^L$& $\bar{S3}_k^L$& $\bar{S4}_k^L$& $\bar{S5}_k^L$ & $\bar{S}_k^L$\\
		& 0.77 & pear & 1.0 & 0.75  & 1.0 & 0.0  & 0.0 & 0.0 & 0.0 \\
		& &  scissors & 0.75 & \textbf{0.25}  & \textbf{0.0} & N/A & N/A & N/A & N/A \\
		& & chips can & 0.5 & 0.5  & 1.0 & 0.48  & 1.0 & 1.0 & 1.48 \\

		& & strawberry & 1.0 & 1.0  & 1.0 &  0.13 & 0.6 & 0.55 & 0.51 \\

		& & tennis ball & 1.0 & 0.75  & 1.0 & 0.07 & 0.4 & 0.4 & 0.43 \\
		& & power drill & \textbf{0.25} & \textbf{0.0}  & \textbf{0.0} &N/A & N/A & N/A & N/A \\
		& & mustard bottle & 0.5 & 0.5  &1.0 & 0.25 & 1.0 & 1.0 & 1.25 \\
		& & medium clamp & 0.75 & \textbf{0.25}  & \textbf{0.0} &  N/A & N/A & N/A & N/A \\
		& & master chef can & 1.0 & 1.0  & \textbf{0.0} & N/A & N/A & N/A & N/A\\
		& & potted meat can & 0.75 & \textbf{0.25}  & 1.0 & N/A & N/A & N/A & N/A \\
		&  & tomato soup can & 0.75 & \textbf{0.25}  & 1.0 & N/A & N/A & N/A & N/A \\
	\vspace{1pt}
	\end{tabular}
	\caption{Results obtained when testing Cardinal Point Grasps~\cite{nguyen2018merging} on the iCub humanoid robot.}
	\label{tab:benchmark-icub}
\end{table*}

\section{CONCLUSIONS}
\label{conclusion}
In this paper we proposed version 1.0 of GRASPA, a benchmarking protocol and a set of metrics to fairly evaluate grasping pipelines tested on diverse robotic platforms. As shown by a practical application (Section \ref{application}), the metrics and the final grasping score we designed allow distinguishing between failures caused by the testing platform and those induced by the limitations of the pipeline itself.

Future work directions for successive releases consist in improving the computation of the grasp quality score $\bar{S3}_L^k$ by allowing users the possibility to specify custom finger joint trajectories during the hand closure simulation and by allowing users to set specific fingertip-object friction coefficients. As outlined in Section \ref{grasp_quality}, version 1.0 of GRASPA employs a grasp quality metric based on analysis of the GWS. This indicator has been shown to be brittle with respect to uncertainty~\cite{kim2013physically}, therefore future development of GRASPA will include a measure of grasp quality that accounts for object dynamics. We also plan to use GRASPA to evaluate the pipeline outlined in \ref{cardinal_point_grasps} and others drawn from the state of the art on a setup equipped with a Franka Panda arm and RGBD cameras in order to compare results. Finally, new objects and layouts can easily be added to the ones presented in this paper to meet the community needs.


\bibliography{bibliography/all_biblio.bib}

\end{document}